# Grasp stability prediction with time series data based on STFT and LSTM

Tao Wang[1,2], and Frank Kirchner[2,3]

*Abstract*— With an increasing demand for robots, robotic grasping will has a more important role in future applications. This paper takes grasp stability prediction as the key technology for grasping and tries to solve the problem with time series data inputs including the force and pressure data. Widely applied to more fields to predict unstable grasping with time series data, algorithms can significantly promote the application of artificial intelligence in traditional industries. This research investigates models that combine short-time Fourier transform (STFT) and long short-term memory (LSTM) and then tested generalizability with dexterous hand and suction cup gripper. The experiments suggest good results for grasp stability prediction with the force data and the generalized results in the pressure data. Among the 4 models, (Data + STFT) & LSTM delivers the best performance. We plan to perform more work on grasp stability prediction, generalize the findings to different types of sensors, and apply the grasp stability prediction in more grasping use cases in real life.

## I. INTRODUCTION

As robotics expands into more applications, grasping has become a necessary ability for robots. In fields like logistics where robots are often required to handle diverse items, grasping is even more pivotal. In order to stably move different items, robots need to have grasp stability prediction to judge the grasping and adjust it to be appropriate [1].

In the area of grasping, sensors of force, torque, and pressure provide feedback with possibly valuable information to judge the grasping state, which is often based on a fixed threshold [2]. Other than the above sensors, more have been invented based on capacity, RGB camera, and time-of-flight (TOF) camera [3]-[7]. However, owing to the existing limitations in data processing and sensor stability, more research is required to study these new sensors' applications in the industrial areas.

At the same time, data of force, torque, and pressure sensors have already been playing an important role in the robotic manipulation processes. We intend to exploit the potential of these various types of sensors by leveraging machine learning methods [8]-[9]. By comparing data from these sensors, we explore algorithms with sufficient generalization to cover all the activities during grasping, which can further promote the extensive applications of these sensors in various industries. In this study, we select the grasp stability prediction case to evaluate the best algorithm.

Grasp stability prediction is critical for advancing functionality in robotics and slip is an important feature of unstable grasping. We use the slip signal to indicate the unstable status when the relative motion is parallel to the contact surface. Predicting the slips before gross motion occurs can ensure a more stable grasping. This predictable incipient slip event can be evaluated with measurements in contact position changes, vibrations, and force changes [10].

As listed above, slip can be identified by monitoring the relative position changes [11]. However, such methods can only judge a slip after it occurs instead of predicting it, while a prediction is much more useful when planning how to safely grasp or lift an object.

To model grasp stability prediction with an optimal method, the sensor must be of adequate quality to achieve the best result as efficiently as possible. We select the force sensor from the above the three measurements because it change faster than the other two when a slip is about to occur, so we can immediately predict the instability based on the changed value [12]. Although some vision-based sensors [13] can judge the stability through a deformation change, which change can only be detected after a big enough force value changes, making these sensors not the optimal choice. Moreover, we expect a more stable and faster prediction, which can be fulfilled by the force sensors with better accuracy and higher frequency.

With data from the force sensor and a method that sets a threshold to judge the instability based on time and frequency field, a reasonable prediction result can be obtained for certain tested items. However, the object's material, weight, and shape may influence the interaction forces, so a large library is required to maintain these threshold values [14]-[15]. Therefore, a learning-based approach may be the most suitable for handling complex situations in the future.

Long short-term memory (LSTM) is a special class of a recurrent neural network (RNN) that can learn long-term dependencies [16]-[17]. This type of network has been applied in speech recognition, translation, and other cases that need to process the time sequence data [18]-[19]. Other researchers have applied LSTM for slip detection, such as convolutional LSTM (ConvLSTM), which performs well in highly dense sensor arrangements [20]. Another research uses the method of connecting the convolutional neural network (CNN) output with the LSTM to process visual-tactile data and has demonstrated good slip detection results [21]. With this strong foundation shown by LSTM methods, we continue to explore these applications with new features to enhance the grasp stability prediction results.

Because unstable grasping always occurs with vibration, the frequency composition can help detect and judge the slip occurrence between the item and the hand surface. Researchers set a threshold for a specific frequency range to judge a slippage [22], and based on which we adopt the Short-time Fourier transform (STFT) [23] to obtain the vibration frequency of the surface area and combine it with the LSTM for more accurate slip prediction results [24].

[1] Dorabot Inc. 518000, Shenzhen, China. [2] Faculty of Mathematics and Computer Science, University of Bremen. [3] Robotics Innovation Center (DFKI RIC), German Research Center for Artificial Intelligence GmbH. E-mail: wangtaofree@gmail.com, tao.wang@dorabot.com.

We develop an algorithm that predicts robotic grasp stability using force data, which can be generalized to other types of time series data. The research begins with analyzing a grasp dataset. Section II introduces related work and our methodology. Section III overviews our experimental setup with different end-effectors. Section IV presents and discusses the experiment results, and Section V offers our conclusions.

## II. METHODOLOGY

Our goal is to predict unstable grasping through the change in force before the relative motion changes.

### A. Grasp dataset

With the requirement of using sufficient force sensor data to test and evaluate the model, we choose a visual-tactile grasp dataset [25]-[26].

The visual-tactile grasp dataset is comprised of 2,550 sets, each of which includes 400 time steps data from 16 force sensors. The force data is acquired by piezo-resistance sensor with a range of 0-10000 mN.

The grasping procedure represented in the dataset is distinguished into four phases of pre-grasp, grasping, lifting, and opening. The force data was acquired during the grasping and lifting phases where some slipping may have occurred during the lifting phase. The dataset records the responses of all the parameters — 10 objects, 3 weights, 3 directions, and 3 grasping forces — as well as tests the generalizability of our algorithm. All the grasping data is labeled with a flag of grasping success or failure, which can be compared with the corresponding video data.

Using this dataset, we can compare different algorithmic model structures and parameters without the hardware setup.

### B. Fourier Transform

During the grasping of an item in a stable state, the force will reach a balance without a significant change. From the relationship between the deformation of a thin, homogeneous plate and the corresponding forces shown in (1) [10], we can know the force is highly related to the deformation of surface $u(x,y)$, when there is a relative motion or obvious deformation, the force $f(x,y)$ can help us judge the slip in a shorter time.

$$\frac{\partial^4 u(x,y)}{\partial x^4} + 2\frac{\partial^2 u(x,y)}{\partial x^2}\frac{\partial^2 u(x,y)}{\partial y^2} + \frac{\partial^4 u(x,y)}{\partial y^4} = \frac{-f(x,y)}{Q} \quad (1)$$

The $u(x,y)$ is the deformation of the central plane of the plate, $f(x,y)$ is the distribution of transverse load along the plate, $Q$ is the parameter related to Young's modulus, Poisson ratio and plate dimension.

As the relationship between force, position and deformation shown in (1), the motion along x and y direction will cause the change in force. We can take the force change on the contact surface as the grasp stability criteria.

Following the idea of judging stability based on force changes, the force data Fourier transform is shown in (2). [27] The change in the joint force $f(t)$ may be similar, component forces $F(w)$ in different frequency can help to show significant difference when slip happened.

$$F(w) = \int_{-\infty}^{\infty} f(t) e^{-jwt} dt \quad (2)$$

We perform a STFT on the force data, as the results are shown in Fig. 1. The red line in the Fig. 1 shows the force change, and the block in the background is the corresponding frequency composition. These significant changes in time and frequency can be detected and classified. We can get a conclusion from the frequency components in Fig. 1, the grasp status can be distinguished with the frequency of 3-5 Hz, and higher frequency resolution and bigger range can be more helpful.

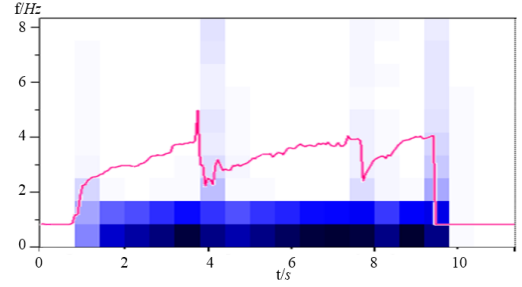

Fig. 1. The use of STFT can help classify instability signals.

### C. Grasp stability prediction framework

With the analysis in frequency domain, we acquire the force data with enough frequency. The force difference $d(t_k)$ between each frame can be shown in (3), the $f(t_k)$ will be influenced by the parameters shown in (1), and related to the time $t_k$. Then we can develop a grasp stability prediction framework both with learning ability and related to time.

$$d(t_k) = f(t_k) - f(t_{k-1}) \quad (3)$$

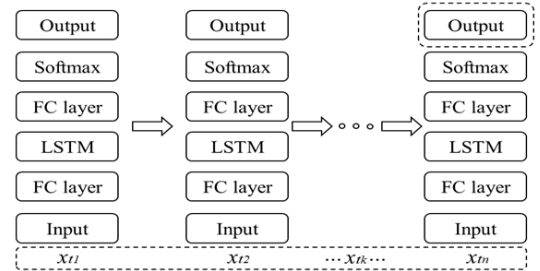

Fig. 2. Structure of the LSTM model.

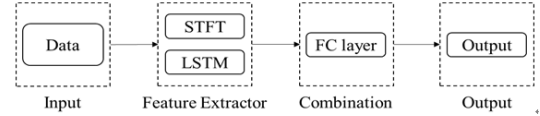

Fig. 3. Grasp stability prediction framework.

As a suitable candidate model, we input a duration of force and STFT data into an LSTM to find the features and output the prediction results. It is expected that the combination of the STFT and LSTM could present a better result. The structure of our LSTM model is outlined in Fig. 2, with inputs from $x_{t1}$ to $x_{tn}$ representing several frames. With the comparison between different output methods, the last step output is taken as the output for the prediction result.

To combine STFT and LSTM, we perform a variety of experiments. Fig. 3 presents our designed framework for grasp stability prediction with an input of the force data followed by a feature extractor that includes the STFT and LSTM components. The subsequent combination of these is a fully connected layer (FC layer), and the output is used as the prediction result.

## III. EXPERIMENTAL SETUP

This work intends to verify the grasp stability prediction performance from our models as well as the generalizability to other types of time series data. Based on these two goals, we developed the following experiments.

### A. Parameters and common configuration

We first define the parameters and common configurations for the experiment to test the framework.

The force data acquiring frequency is 16.7 Hz, the target slipping frequency is 3-5 Hz, and we set STFT with 20 time steps window size and use a rectangular window. We shift the windows with every time step, resulting in data size of n×10 (e.g., n steps with 10 frequency bands). Because the time from the grasping phase to the opening phase is shorter than 9 seconds, we choose 160 as the batch size to guarantee enough effective data.

The label in the datasets only represent the grasping result, so we must label the slip time manually. Based on the time steps of the robotic hand grasping, lifting, and releasing the item, we define that the item will drop when the force is at 0. Because slippage must occur before a drop, we label the slip time with the assumption that the real slip time will happened during 20 time steps (1.2s) before the drop, and the actual success rate of label should be lower than 93.75% (160 steps in total and estimated with average distribution).

In this experiment, we built the LSTM model with Tensorflow [28]. With some primary tests, we chose the basic LSTM core with 128 as the unit number. Consider the grasp stability prediction as a 2-type classification problem, we produced an output with softmax. Then we initialized the model with zeros, used cross-entropy as the loss function. For the hyper-parameters optimization, we tuning them based on the comparison of output result. Learning rate is more important in the training process, and we set the Adam optimizer [29] with a learning rate of 0.0006 during the training process.

### B. Grasp stability prediction evaluation on the dataset

The grasp stability prediction dataset combined 10 objects with 16 channels of force data. The following key issues must be resolved before we start the experiment:

1. How does the LSTM perform compared to the other solutions?
2. How to utilize the dataset with multiple channels, directions, and items?
3. How to compare the possible combinations of the STFT and LSTM algorithms?

To consider these issues, we refer to the paper that introduced the dataset.

With the first issue, we treat grasp stability prediction as a classification problem, and use 16 channels of data as input to perform the classification. We compare algorithms of NB [30], KNN [31], and SVM [32] with the same dataset. Because the prediction result of the whole grasping process is important, we prefer to use the prediction success rate rather than the $F_{score}$ used in some classification issues.[9] Two criteria are set to evaluate the solutions: (a) the success rate, or the rate that the prediction matches with the label; (b) the ahead drop rate, or the rate that the prediction is ahead of the step which the item is dropped. A comparison of these criteria is listed in Table I.

TABLE I COMPARISON AMONG DIFFERENT SOLUTIONS [25]

| Model | Success rate | Ahead drop |
|---|---|---|
| NB | 68.81% | 65.69% |
| KNN (k=3) | 79.70% | 81.76% |
| SVM | 84.67% | 66.67% |
| LSTM | 84.60% | 85.88% |

According to the table, the LSTM approach has a high success rate over 84% and demonstrates the best performance in the ahead drop rate, which suggests these results can be used as approaches for prediction instead of detection.

The second issue matters to our goal of generalization, as shown in Fig. 4, different directions and items will influence the touchpoints position during grasping. Such as training with data of back direction, it is expected that the model would have poor performance along the top direction. As Table II suggests that the channel difference plays an important role, and we will only use the single channels as the input to decrease its influence for a more generalized solution. Therefore, each set of data input into the network will only include one channel.

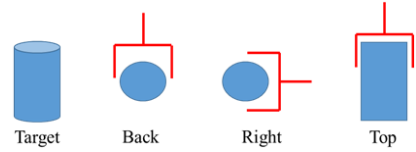

Fig. 4. Grasp from different directions.

TABLE II COMPARISON AMONG DIFFERENT DIRECTIONS [25]

| Success rate | Back (test) | Right (test) | Top (test) |
|---|---|---|---|
| Back (train) | 87.38% | 77.36% | 76.91% |
| Right (train) | 78.56% | 82.94% | 76.15% |
| Top (train) | 71.42% | 66.94% | 83.90% |

For the third issue, we need to explore suitable network structures with the combination of LSTM and STFT. A variety of designed frameworks are shown in Fig. 5.

In option (a), we input the force data into LSTM to obtain the output results. Option (b) takes the frequency data as input, and option (c) uses the ten channels frequency data and one channel force data as input. Option (d) builds two LSTM parts to separately input the frequency data and the force data, and then combines the output.

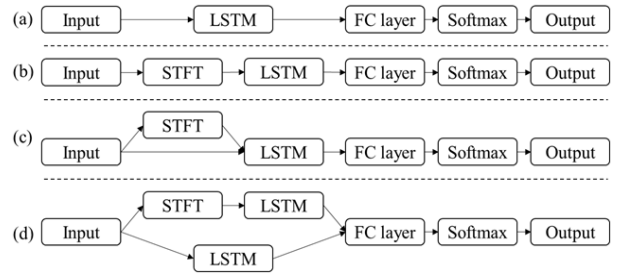

Fig. 5. Different framework options of LSTM, STFT & LSTM, (Data + STFT) & LSTM, and LSTM + STFT & LSTM.

### C. Verification on real situations

Here, we verify the models with the DoraHand dexterous hand in Fig. 6 (a). [33] The force sensor in each finger is similar to the dexterous hand's sensor [26] in the dataset with the data frequency at 40 Hz. The comparison of different models' performance is based on the sensor data acquired from the DoraHand's picking process. What's more, we also try to verify the grasp stability prediction function in real situations, where 4 different methods were defined as follows:

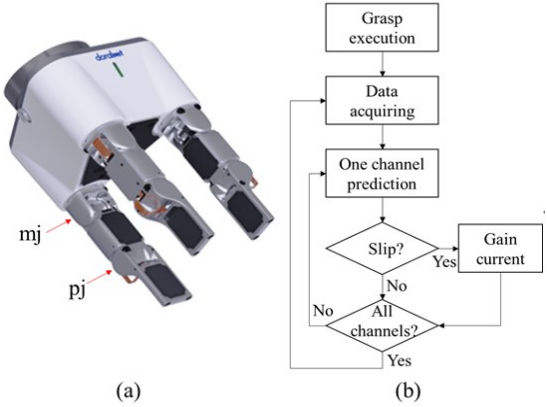

Fig. 6. DoraHand three-figner dexterous hand and grasping strategy.

1. Grasp a heavy item with an initial light force;
2. Grasp a light item which suddenly gains weight;
3. Gradually adding weight to the item;
4. Rotate the item inside hand;

The 4 methods represent four different grasping situations in our daily life: grasping an item with its weight unknown; grasping an item while taking a sudden impact; putting something else into the item; grasping something with low friction.

As the grasping strategy in Fig. 6 (b) shows, the current of pj and mj joint will be increased by 5 mA and 10 mA separately after the grasp stability predicted. The calculation time of prediction is around 4 ms for each sensor, so with 16 sensors in the hand the frequency of prediction is around 16 Hz (64ms in total).

The initial current for mj and pj are 25 mA and 50 mA. The joint current for 2 kg payload grasping are around 100 mA and 200 mA, which means the slipping should be detected for over 15 times during the process.

### D. Test generalizability

In this experiment, we verify the generalizability with the pressure sensor data. The pressure sensor data was measured from the suction cup gripper that is widely installed in the industry grasping cases, as seen in Fig. 7 (a). With an algorithm to predict the grasping stability, the robot can adjust the velocity and acceleration to guarantee the grasping performance.

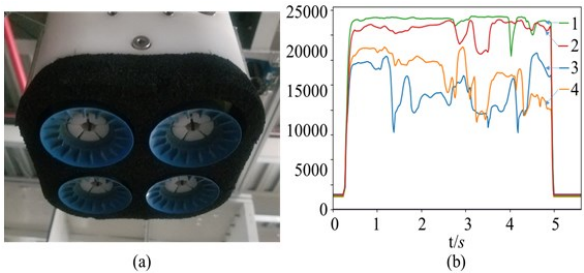

Fig. 7. Gripper with suction cups and data from the pressure sensor.

We collect the pressure sensor data with 71 Hz when the robot performs a pick & place task. When the gripper cannot provide enough suction force to hold the item stable, the pressure in the suction cup will present an obvious change, like when the item deforms or an air leak occurs due to a relative motion change.

Fig. 7 (b) shows the change in air pressure during one pick & place task. The 4 channels represent 4 independent pressure sensors, each with a non-pressure value of around 6400, the preset zero position (the data is 16 bit digital data in a range of 0-65536, and the initial value are 6458, 6263, 6357, and 6458). Because the pressure unit does not influence the result, we did not calibrate the relationship between the analog signal and the air pressure.

## IV. RESULTS AND DISCUSSION

The following results were obtained from the experimental setups described in the previous section.

### A. Comparison of multiple models

As shown in Fig. 5, we designed four frameworks for the comparison and the results are listed in Table III. The (Data + STFT) & LSTM model demonstrates the best success rate, and LSTM offers the best performance in the ahead drop rate.

TABLE III COMPARISON AMONG DIFFERENT MODELS

| Model | Success rate | Ahead drop |
|---|---|---|
| LSTM | 75.27% | **84.63%** |
| STFT & LSTM | 78.39% | 67.88% |
| **(Data + STFT) & LSTM** | **80.09%** | 67.50% |
| LSTM + STFT & LSTM | 78.93% | 69.25% |

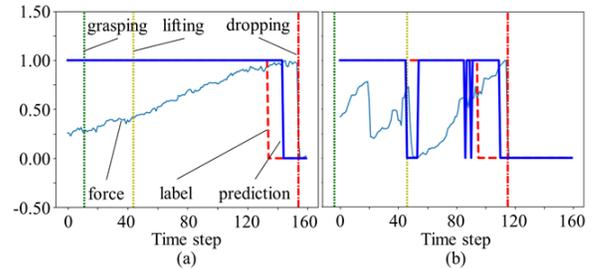

Fig. 8. Prediction result by (Data + STFT) & LSTM.

Predictions from the (Data + STFT) & LSTM framework are shown in Fig. 8 (a) and (b). The vertical axis stands for the Boolean output of stable or not (1 means stable, 0 means unstable), and the horizontal axis represents the time step with data frequency at 16.7 Hz. The green dotted vertical line in the left marks the starting time of the grasping, the yellow dotted vertical line represents the starting time of lifting the item, and the red dash-dot vertical line shows the dropping time. The blue thin line represents the normalized force value, the red dashed bold line illustrates the labeled data, and the blue bold line is the prediction result.

In Fig. 8 (a), the force changes a lot after the lifting time step, and the slippage is hard to judge from the force change. The blue bold line (prediction) shows the same result as the red dashed bold line (labeled data) in most time and the predicted slipping point is ahead of the dropping time. Because the slipping time is labelled 20 time steps before the dropping time and the slipping time can't be shown precisely in the graph, the prediction data can't reach 100% result and this type of prediction can be regarded as the good result.

Fig. 8 (b) shows an obvious slip prediction at the lifting time step. As the force changes to 0, we can assume the occurrence of a slip. However, after checking the video, there was no obvious evidence of slipping. This "wrong prediction" is a result of the single channel input. At that time, the robotic hand lifted the item so the forces of different fingers change and then reached a new balance to prevent slipping. Only this channel witnessed this type of "slip", but it still should be stable status.

Based on these partial results, the following three points are suggested.
1. The STFT data improves the success rate and enables a more precise and stable prediction.
2. The time data directly improves the ahead drop rate, and it is crucial for the grasp stability prediction.

Training with a single channel performs a lower success rate (compared with 84.60% in Table I), possibly because data from multiple channels can provide additional information, such as the contact information between hand and item. Therefore, we can combine different data channels in specific use cases to improve the success rate.

### B. Verification on dexterous hand

For the comparison on DoraHand, the performance of 4 different models is shown in Table IV and Fig. 9. The success rate is similar to the result of the dataset, LSTM and (Data + STFT) & LSTM show a good performance at over 85%. And the low ahead drop rate may be caused by different data acquiring methods. We tested the grasp stability prediction in a relatively real situation, where no slipping happens in the most testing time, so the ahead drop rate is at 0 in these cases.

TABLE IV COMPARISON ON DORAHAND DATA

| Model | Success rate | Ahead drop |
|---|---|---|
| LSTM | 89.38% | 7.34% |
| STFT & LSTM | 80.19% | 24.68% |
| (Data + STFT) & LSTM | 87.24% | 9.67% |
| LSTM + STFT & LSTM | 74.39% | 32.53% |

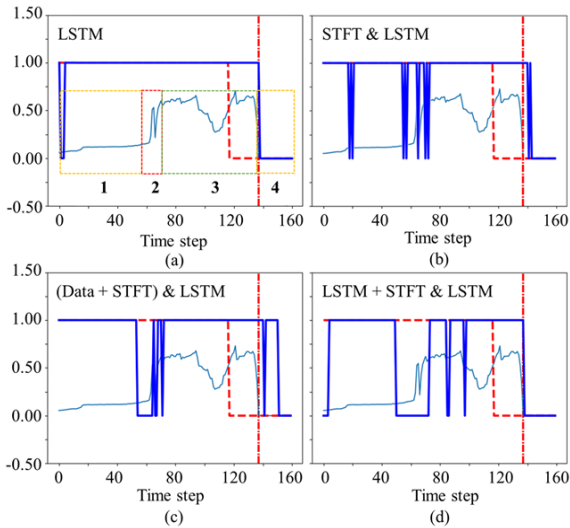

Fig. 9. Prediction results by different models on dexterous hand data.

Comparing the details of the prediction results, we labeled 4 different phases in Fig. 9 (a), (1) adding force, (2) slipping, (3) holding the item, and (4) dropping the item.

LSTM shows the best success rate like Fig. 9 (a), but the prediction result only showing the dropping time in phase 4 can't meet real application requirement. The other three models with STFT input predict the slipping result in phase 2 and output the stable result in phase 3, which shows the grasp stability prediction capability and tells the difference of the force change between phase 2 and 3. Such comparison shows that the frequency data can help to predict the slippage through different force value changes.

For the verification on grasp stability prediction performance, we designed the robot to lift the item with light force and tested it in different situations, as shown in Fig. 10, experiment (a) and (b) were separately tested with both disable and enable prediction function, the (Data + STFT) & LSTM was tested.

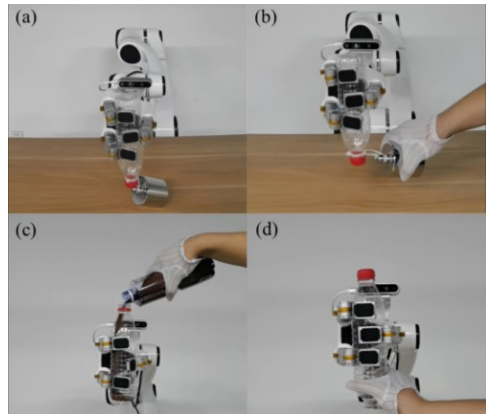

Fig. 10. Test with 4 different situations.

The comparison of disabled and enabled prediction function of the 1st and 2nd experiments is shown in Fig. 11. The item is dropped like in Fig. 11 (a) and Fig. 11 (c) when we disabled the slip prediction function; and it is lifted like in Fig. 11 (b) and Fig. 11 (d) when we enabled the slip prediction function. It shows that the slip prediction algorithm can react fast enough to hold the item even if the item has a sudden weight increase of 2 kg. The 3rd grasping in Fig. 10 (c) become more and more stable with the weight adding. The 4th grasping in Fig. 10 (d) become tighter after several rotations.

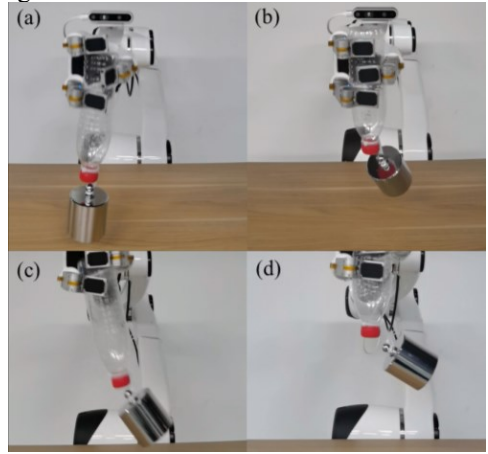

Fig. 11. Comparison between disable and enable slip prediction.

We record the force feedback during these experiments, we can easy find that the force increases with the prediction result, as shown in Fig. 12. Fig. 12 (a) is the case when the robot lifts an item while taking a 2 kg impact, through which we can see that the force increases its speed within 0.15 s, and the force drop just happened between 0.925-1.05 s. This means that the algorithm is fast enough to predict the stability within 1-2 frame data (1 time frame include 16 channels and takes 64 ms for 16 times prediction). Fig. 12 (b) is the case when the robot rotates the item inside the hand, showing that the force increases step by step after each prediction.

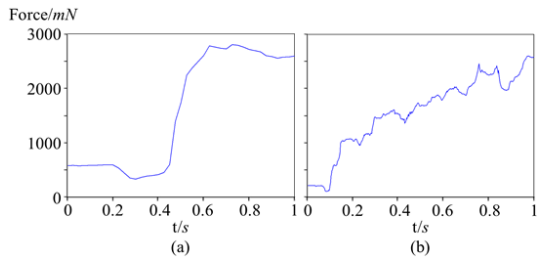

Fig. 12. Force increases with slip prediction result on different test cases.

With this verification applied on the dexterous hand, we can sum up the following key points:
1. Frequency data can help to find more slip features.
2. Labeling method is critical to the training process. We need a new dataset to achieve more accurate results, labeling which stage matters to the use cases. Labeling methods with other sensors like the accelerometer may be helpful [34].
3. The current model based on STFT and LSTM can be applied in real grasping case which already tested by the DoraHand.

*C. Generalization to different situations*

For our experiments, we selected data from the pressure sensors as the input and separated the continuous data into 316 batches, each with a size of 160 time steps. The results are shown in Table V, Fig. 13 (a) and (b).

TABLE V COMPARISON ON PRESSURE DATA

| Model | 17.75 Hz | | 71 Hz | |
| --- | --- | --- | --- | --- |
| | Success rate | Ahead drop | Success rate | Ahead drop |
| LSTM | 64.15% | 6.94% | 78.69% | 4.11% |
| STFT & LSTM | 57.11% | 16.67% | 71.91% | 18.67% |
| (Data + STFT) & LSTM | 58.25% | 6.94% | **78.80%** | 1.90% |
| LSTM + STFT & LSTM | 60.11% | 12.50% | 76.65% | 8.23% |

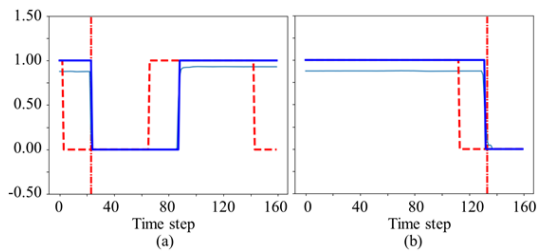

Fig. 13. Prediction results by LSTM on 17.75 Hz and LSTM + STFT & LSTM on 71 Hz data.

Among the four models, (Data + STFT) & LSTM performs the best with the highest success rate in the 71 Hz test, showing a good generalization ability. And all these models have a lower performance in the 17.75 Hz test, the reason may be the dropping process of the suction cup gripper is much faster than the dexterous hand. The low ahead drop rate in all these experiments may also attribute to the low data frequency which cannot support to predict the stability.

As seen in Fig. 13 (a), the output prediction result (blue bold line) nearly superimposes with the labeled pressure data, but the pressure changes too fast to make a prediction ahead of the drop. The pressure value drops rapidly when there is a significant vibration or air leak. The 20 steps ahead of the drop for slip label is too many in this case, making the success rate decrease to 62.5%.

Fig. 13 (b) shows a good prediction result by LSTM + STFT & LSTM on 71 Hz data, the prediction result locates between the unstable label and the drop label, but the data still changes too fast and not enough for prediction. A model trained by higher frequency pressure data and better label methods may help to achieve a better result.

From this experiment, we sum up the following key points:
1. (Data + STFT) & LSTM show a good capability in the stability prediction and can be generalized to other data.
2. Features in the time series are easier to transfer to other targets than those in the frequency domain.
3. Higher frequency data can help to improve the training and prediction result.

V. CONCLUSION

We introduced the requirements for grasp stability prediction using force sensor data with the possibility to generalize it to time series data. We proposed a grasp stability prediction solution with STFT and LSTM that combines the information from time and frequency domain.

This concept was based on a published dataset and verified with new sensor data measurements. A comparison of several potential models suggests that combining STFT and LSTM improves the success rate, and the input data with a higher frequency is significantly beneficial to improve the results. The experiment on the DoraHand dexterous hand verified our training results and showed the capability of grasp stability prediction model in real use case. Tests on the pressure sensor data demonstrated that the solution's generalizability is good enough. The data frequency significant influences the result when using the STFT. The time domain performs better than the frequency domain in generalizability.

Following our discussion, the model of (Data + STFT) & LSTM improves the precision of grasp stability prediction, and can continue to be improved in term of ahead drop rate. In the future, we will explore more STFT & LSTM-based solutions for grasp stability prediction, apply these candidate models to more real situations, and continue to generalize the findings of time series data with enhanced performance to more applications.